%% file: GaitPattern.tex
\documentclass[a4paper, 10pt, conference]{ieeeconf}      % Use this line for a4
\usepackage{FG2023}
\usepackage{amsmath}
\usepackage{graphicx}
\usepackage{gensymb}
\usepackage{xcolor}
\usepackage{subcaption}
\input{math_commands.tex}

\usepackage{multirow}
\usepackage{array}
\usepackage[T1]{fontenc}
\usepackage[utf8]{inputenc}
\usepackage{authblk}
\renewcommand{\thefootnote}{\fnsymbol{footnote}}
\FGfinalcopy % *** Uncomment this line for the final submission

\IEEEoverridecommandlockouts                              % This command is only
\def\FGPaperID{34} % *** Enter the FG2023 Paper ID here

\title{\LARGE \bf
Multi-Modal Human Authentication Using Silhouettes, Gait and RGB
}

\author{\parbox{16cm}{\centering
    {\large Yuxiang Guo$^*$, Cheng Peng$^*$, Chun Pong Lau, Rama Chellappa}\\
    {\normalsize
    Johns Hopkins University, Baltimore, MD, USA\\}}
}

\begin{document}
\ifFGfinal
\thispagestyle{empty}
\pagestyle{empty}
\else
\author{Anonymous FG2023 submission\\ Paper ID \FGPaperID \\}
\pagestyle{plain}
\fi
\maketitle

%%%%%%%%%%%%%%%%%%%%%%%%%%%%%%%%%%%%%%%%%%%%%%%%%%%%%%%%%%%%%%%%%%%%%%%%%%%%%%%%
\input{FG2023/abstract}
\input{FG2023/introduction}

\input{FG2023/related_work}

% \input{FG2023/GaitPattern}
\input{FG2023/method}
\input{FG2023/experiments}

\input{FG2023/conclusion}
\input{FG2023/acknowledge}

%%%%%%%%%%%%%%%%%%%%%%%%%%%%%%%%%%%%%%%%%%%%%%%%%%%%%%%%%%%%%%%%%%%%%%%%%%%%%%%%
%\newpage
{\small
\bibliographystyle{ieee}
\bibliography{egbib}
}

\end{document}

%% file: math_commands.tex
\usepackage{amsmath,amsfonts,bm, bbm}

\def\eqref#1{equation~\ref{#1}}
\def\1{\bm{1}}

\def\vf{{\bm{f}}}

\def\vs{{\bm{s}}}

\def\mI{{\bm{I}}}

\def\mS{{\bm{S}}}

\def\mV{{\bm{V}}}

\def\mX{{\bm{X}}}

\DeclareMathAlphabet{\mathsfit}{\encodingdefault}{\sfdefault}{m}{sl}
\SetMathAlphabet{\mathsfit}{bold}{\encodingdefault}{\sfdefault}{bx}{n}

\def\gD{{\mathcal{D}}}

\def\gF{{\mathcal{F}}}

\def\gS{{\mathcal{S}}}

\def\gV{{\mathcal{V}}}

\def\gY{{\mathcal{Y}}}

%% file: FG2023/abstract.tex
\begin{abstract}
    \footnotetext[1]{Equal contribution.}Whole-body-based human authentication is a promising approach for remote biometrics scenarios. Current literature focuses on either body recognition based on RGB images or gait recognition based on body shapes and walking patterns; both have their advantages and drawbacks. In this work, we propose Dual-Modal Ensemble (DME), which combines both RGB and silhouette data to achieve more robust performances for indoor and outdoor whole-body based recognition. Within DME, we propose GaitPattern, which is inspired by the double helical gait pattern used in traditional gait analysis. The GaitPattern contributes to robust identification performance over a large range of viewing angles. Extensive experimental results on the CASIA-B dataset demonstrate that the proposed method outperforms state-of-the-art recognition systems. We also provide experimental results using the newly collected BRIAR dataset.
\end{abstract}

%    Whole-body recognition provides the promise of more robust biometric identification in the wild due to a larger target size. Current literature focuses on either body recognition based on RGB images or gait recognition based on body shapes and walking patterns; both have their merits and drawbacks. In this work, we propose Dual-Modeal Ensemble (DME), which combines both the RGB and silhouette modality to achieve more robust performances for indoor and outdoor recognition. Within DME, we proposes GaitPattern. GaitPattern is inspired by the Frieze gait Pattern found in traditional gait recognition literature and addresses the challenges in indoor gait recognition over different viewing angles. Extensive experiments demonstrate that our proposed method outperforms current state-of-the-art recognition systems on both public and privatedly collected, outdoor recognition datasets.

%% file: FG2023/introduction.tex
\section{Introduction}
\renewcommand{\thefootnote}{\fnsymbol{footnote}}

Body recognition from videos is an important yet challenging computer vision task. The objective is to determine whether the subjects in different videos have the same identity. Similar to face recognition, body recognition has applications ranging from surveillance to intelligent transportation. In particular, body recognition is more robust than face recognition in many unconstrained situations, where faces are acquired in non-cooperative conditions. 
Gait recognition~\cite{fan2020gaitpart,lin2021gait,chao2019gaitset}, which recognizes people based on their walking patterns over time, similarly performs recognition based on the body, but typically uses human silhouettes as input. In comparison, body recognition methods exploit silhouettes as well as RGB data.

There are advantages and disadvantages when body or gait is used for remote identification separately. Specifically, RGB videos contain rich information that can lead to significantly better performance; however, such information can be ineffective in situations involving clothing change and image quality degradation. The silhouette modality allows a recognition network to focus purely on the subject's body shape and gait; as such it is less susceptible to clothing changes. This robustness comes at the cost of generally lower performance compared to methods that exploit RGB data. Motivated by this observation, we propose Dual-Modal Ensemble (DME), which performs learning in both RGB and silhouette domains and leverages the model ensemble to extract the most robust features for unconstrained situations. DME is also flexible and can be used separately when the image condition is ill-suited for an individual modality. 

Furthermore, within DME, we address several issues in the \emph{gait recognition} space. Due to the high computational cost of processing video data through neural networks,  State-of-The-Art (SoTA) methods rely on temporal pooling operations to reduce the dimension from 3D to 2D. As such, spatial information plays a major role in these recognition algorithms. It has been observed across the gait recognition literature that such systems tend to generalize poorly over different view angles~\cite{fan2020gaitpart,lin2021gait,chao2019gaitset}. An efficient gait representation in the temporal domain is thus highly desired and can complement conventional gait recognition systems. Such a temporal representation can provide knowledge of the camera's viewing angle, assuming the cameras are stationary and the subject is walking in a normal pattern.

Inspired by traditional gait recognition approaches that produce such a temporal representation through a Double Helical Signature (DHS)~\cite{ran2010applications}, we propose GaitPattern, learned using a deep neural network and incorporate it into the recognition system. We empirically find that the introduction of GaitPattern significantly improves performances at challenging view angles.

% While gait recognition offers certain advantages, the silhouette inputs discard the rich information contained in the color space and can be more severely impacted by other conditions, e.g. incomplete observation of the body. To leverage the advantages in both gait and RGB-based body recognition, we propose an Dual-Modal Ensemble system, which combines RGB and silhouette features together as an overall identification feature. While the human silhouette is theoretically contained in the color space, the independent learning in each modality allows respective networks to focus on distinct features. As such, the ensemble recognition system significantly outperforms a color-space only recognition system with comparable network depth. Such a system is also flexible, and can be used separately when the image condition is ill-suited for an individual modality. 

%Most body recognition works so far are designed evaluated in controlled, indoor settings~\cite{gaitgl,c1,fillintheblank}, which do not reflect challenges such as image distortion due to turbulence or other atmospheric conditions, complicated human poses, and low image resolutions. This issue is particular severe when the gallery images are high quality, while probe images are low quality.

In summary, this paper makes the following contributions:

\begin{itemize}
    \item We propose Dual-Modal Ensemble, which incorporates both gait and RGB modalities to perform body recognition; such a design allows for both flexibility and superior performance both for indoor and outdoor scenes.
    \item We propose GaitPattern, which efficiently incorporates temporal information into gait recognition and leads to performance improvements in indoor scenes, especially for challenging viewing angles.
    \item We perform extensive evaluations on both CASIA-B and Biometric Recognition and Identification at Altitude and Range (BRIAR) dataset, an unconstrained face and body recognition dataset.
    
\end{itemize}

%% file: FG2023/related_work.tex
\section{Related Work}
\label{related_work}

\subsection{Gait Recognition}
% Gait recognition is a way to identify human subjects by their walking pattern. In general gait representation could b\
% e described as 2D and 3D. For 3D, there are many works \cite{zhao20063d, bodor2009view, ariyanto2011model} using 3D g\
% ait representation capturing by multiple cameras. 3D gait representation contains rich information and give a compell\
% ing performance for gait representation. However, it is usually not practical to have multiple cameras to obtain the \
% 3D model. Moreover, it is more computational costly than 2D gait. To represent 2D gait, there are two typical approac\
% hes. One is using gait silhouettes, which are the binary segmentation masks of the subjects. Another is skeleton-base\
% d \cite{an2020performance, liao2020model}, which uses the keypoints of the joint and extract features from the skelet\
% on. In this work, we focus on 2D gait with silhouette representation.

Gait recognition aims to identify human subjects by their walking pattern. In general, gait representation can be done in 2D and 3D. There are many works \cite{zhao20063d, bodor2009view, ariyanto2011model} that use 3D gait representations, which are obtained from multi-camera setups. While 3D gait representation contains rich information and provides compelling performances, a multi-camera setup is often impractical for general applications. Moreover, such a 3D approach is computationaly costlier than 2D approaches. There are typically three approaches to obtain 2D gait representation. One approach uses gait silhouettes~\cite{fan2020gaitpart,lin2021gait,chao2019gaitset,hou2020gait,wu2016comprehensive,huang20213d,kale2002gait,kale2004identification,liu2006improved,han2005individual}, extracted from video frames of walking subjects. The  skeleton-based approaches~\cite{an2020performance, liao2020model, teepe2021gaitgraph,teepe2022towards} use the key-points of the joints, i.e. a skeleton representation, and extract features from the skeleton. Another approach fuses features from silhouettes and skeleton~\cite{wang2022two,peng2021learning}.

\begin{figure*}
    \centering
    \includegraphics[width=\textwidth,height=.24\textheight]{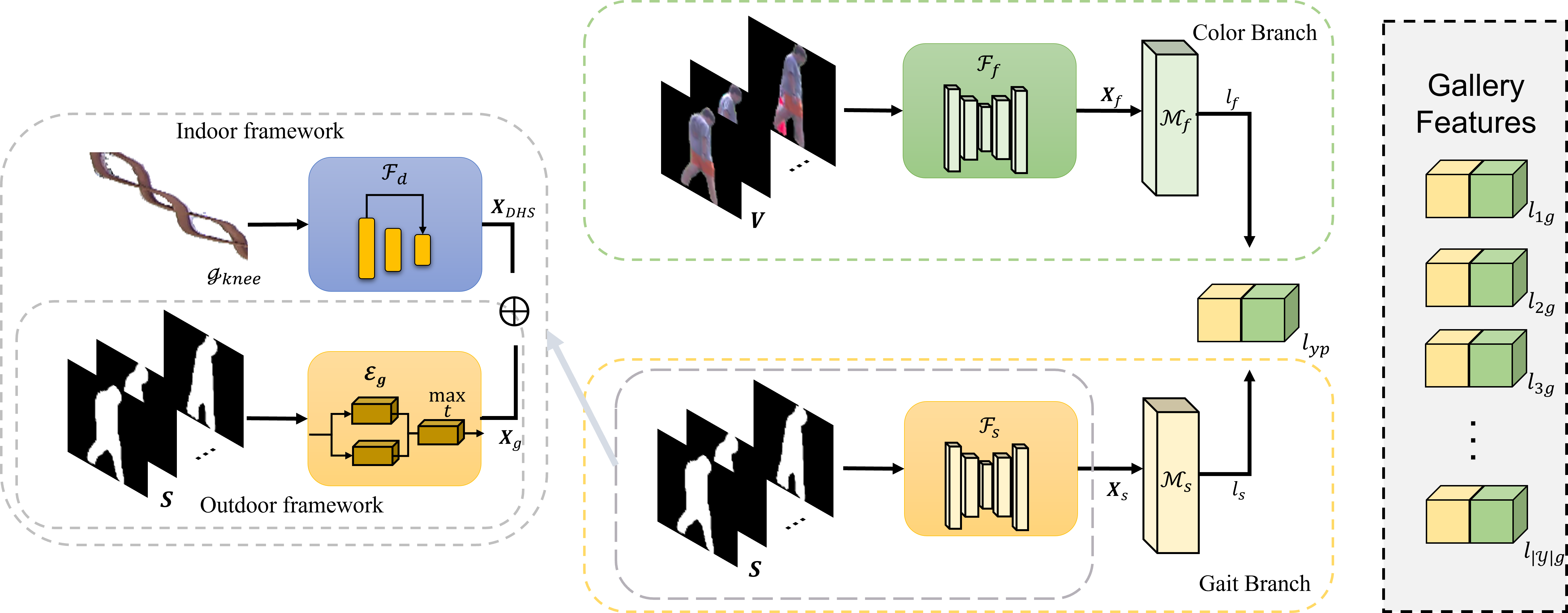}
    \caption{The overall pipeline of the proposed method. The method is divided into gait and RGB two branches. They both use backbone convolutions $\mathcal{F}_{f}$ and $\mathcal{F}_{s}$ following the multiple-layer perceptrons (MLP) $\mathcal{M}_{f}$ and $\mathcal{M}_{s}$ to generate the feature $l_y$. The features are extracted from the gallery ($l_{yg}$) and probe ($l_{yp}$) videos to measure the distance between them. For gait branch, the DHS feature $X_{DHS}$ will be concatenated with the that of gait $X_{g}$, if the DHS feature is available.}
    \label{fig:pipeline}
\end{figure*}

\subsection{Double Helical Signature}
\label{subsec:DHS}

% Frieze Pattern is one of the essential tools to analysis a pair of dual dimensional sequence with a periodic along time axis, which is the nature of the human walking. So, 

Human gait represented as a Double Helical Signature (DHS)~\cite{ran2010applications}, also known as a type of ``Frieze pattern"~\cite{niyogi1994analyzing}, has been proposed and applied for gait sequence analysis in the past, when limited computational power encouraged researchers to find efficient representations to perform recognition. As shown in Fig.~\ref{fig:sub1}, DHS can be generated by extracting the horizontal slices of the subject's knee from every frame in a video and stacking the slices to form a time-width image. Prior works~\cite{ran2010applications,niyogi1994analyzing,liu2002gait, liu2004computational} have shown that DHS can efficiently encode parameters such as step size and gait/walking rate and also determine if the subject is carrying objects.

\subsection{Video-based Body Recognition}
Due to the availability of rich spacial and temporal information and computational resources, videos can be more effective for whole body recognition. Accurate capture of temporal information is a primary challenge. There are mainly two methods to capture the temporal relations. Temporal attention~\cite{fu2019sta,zhou2017see,zhao2019attribute} is one method, which aims to determine the effectiveness of each frame and discards the low-quality ones. The other method is self-attention or Graph Convolutional Networks (GCNs)~\cite{li2019global,yan2020learning,bai2022salient}, which exploits the temporal relations.
% The other methods including optical flow~\cite{chen2018video}, 3D CNNs~\cite{li2019multi}, and the performance is restricted due to high computational cost and limited temporal resolution.

\subsection{Remote Biometric Recognition in the Wild}
Remote biometrics in the wild has been studied for over fifteen years. Some of the early efforts in remote biometrics recognition were based on gait and faces. As discussed before, the gait recognition works reported in~\cite{han2005individual,kale2002gait,kale2004identification,liu2006improved} and video-based recognition works reported in~\cite{zhou2002probabilistic,zhou2004visual} are some examples of early works on remote biometric recognition efforts, although the images and videos were collected at varying degrees of complexity. More recently, face recognition at distances of 300-1000 meters has been addressed in~\cite{lau2020atfacegan,lau2021atfacegan,yasarla2020learning,yasarla2021learning}. For example,~\cite{lau2020atfacegan,lau2021atfacegan} propose a generative single frame restoration algorithm which disentangles the blur and deformation due to atmospheric turbulence and reconstructs a restored image. Extensive experiments demonstrate the effectiveness of the proposed restoration algorithm,
which achieves satisfactory results at 300 and 650 meters, but the low-accuracy detection of faces at 1000 meters affects the overall performance. Similar efforts using generative models have been reported in~\cite{yasarla2020learning,yasarla2021learning}. While many datasets such as CASIA-B and CASIA-E have been collected and many traditional and deep learning-based algorithms have been evaluated on these datasets, these datasets also vary in terms of how unconstrained the collected videos are. Recently, the IARPA BRIAR program has collected datasets of faces and whole bodies at distances upto 500 meters. We present the results of whole-body recognition based on silhouettes, gait and RGB using the BRIAR dataset.
% \subsection{Biometric Recognition in the Wild}
% While many works exist for whole-body biometric recognition, these works are generally done in controlled environments, e.g. indoor and with precise segmentation labels. Significant challenges exist when these recognition systems are deployed in the wild, e.g. in outdoor situations. In particular, the images or videos can be severely degraded by atmospheric turbulence, which consisting of space-varying blur, deformation and noise. These degradation affects our ability to obtain human segmentation masks and to perform gait recognition.
% There are some works in face recognition that focuses on restoring turbulence-degraded face images \cite{lau2020atfacegan, lau2021atfacegan, yasarla2022cnn} and locating facial key-points \cite{lau2021semi}; however, there has not been much exploration for body and gait recognition in outdoor situations and at a distance,  %This motivates us to propose a new method that is robust in this tough situation. 

%% file: FG2023/method.tex
\section{Method}

% In this section, we introduce a novel model integrated with Dual-Modal Ensemble to fuse the GaitPattern, gait and color in a complementary manner, as shown in Fig.\ref{fig:pipeline}.  The system is mainly divided into gait and color two branches. They both use backbone convolutions following the multiple-layer perceptions (MLP) to generate the feature. The features are extracted from the gallery and probe videos to measure the distance between them. For gait branch, the DHS feature is concatenated with the that of silhouette when the video is taken indoor (constrained). Otherwise (unconstrained), only the silhouette feature is used, since the key-point estimation is likely accurate in constrained environment, which makes DHS generated reliably.

\subsection{Problem Formulation}\label{PF}

Suppose the target video $\mV = [\vf_1, \vf_2, \dots, \vf_T] \in \gF := \mathbb{R}^{H \times W \times T \times 3}$ consists of $T$ video frames $\vf_i$, where $H$ and $W$ are the height and width of the frame. In $\mV$, we assume there is a corresponding subject, labeled as $y \in \gY$, where $y \in \{ 1, 2, \dots, |\gY| \}$ and $\gY$ is the dataset. To represent gait, we obtain the human silhouette sequence $\mS = [\vs_1, \vs_2, \dots, \vs_T] \in \gS := \{0, 1\}^{H \times W \times T \times 1}$, which consists of binary masks obtained by image segmentation. Let $F_\theta$ be a parameterized model which maps any video in $\gF$ to a feature vector $\mX$ in $F_\theta(\gV)$. An accurate whole-body recognition system can map two videos $\mV_1$ and $\mV_2$ with the same identity to features that are close in a certain feature space, i.e. $\gD(F_\theta(\mV_1), F_\theta(\mV_2))$, where $\gD$ is an appropriate distance function, e.g. Euclidean. Specifically, this can be done by using either RGB frames, silhouette sequences, or both. For body recognition using RGB frames, the videos are first preprocessed by masking out the background based one $\mS$ to prevent overfitting, i.e. $\mV = \mV_{\textrm{orig}} \odot \mS$, where $\mV_{\textrm{orig}}$ is the original video and $\odot$ is the Hadamard product.

In the following sections, we describe DME, which uses RGB and silhouette together for more robust performances, and \emph{GaitPattern}, a gait recognition method that leverages the efficient Double Helical Signature representation ~\cite{ran2010applications}.

% In the following sections, we describe \emph{GaitPattern}, a gait recognition method that leverages the efficient Double Helical Signature (DHS)~\cite{}, and Dual-Modal Ensemble that uses body the RGB and silhouette modality for more robust performances.

\subsection{Dual-Modal Ensemble}
\label{DME}

% While gait is a powerful feature to perform body recognition, it  loses the rich semantic information present in RGB videos. This loss of information is especially significant in unconstrained situations, where the silhouettes are neither provided nor can be precisely extracted. RGB information, on the other hand, is more prone to degradation by the acquisition environment, e.g. camera distance, weather effects, change of clothes, etc. The proposed method leverages the strengths of both gait and RGB modalities while suppressing their weaknesses. 

Dual-Modal Ensemble proposes to use two CNN architectures to extract features from individual modalities. As shown in Fig.~\ref{fig:pipeline}, this process can be expressed as:

%. since the background information in RGB video is not a desired robust feature, we firstly mask out the background information by using the silhouette sequence

\begin{equation}
    X_{f} = \mathcal{F}_{f}(\mV),\\
    X_{s} = \mathcal{F}_{s}(\mS),
\end{equation}

\noindent where $\mathcal{F}_{f}$ and $\mathcal{F}_{s}$ are the feature extraction CNNs for the respective modalities. Generally, $\mathcal{F}_{f}$ and $\mathcal{F}_{s}$ can be any well-performing CNN architecture, and we elaborate the details of network design in Section~\ref{GFE} and~\ref{rgbfeat}. Using separate feature extraction blocks for each modality allows for better flexibility, as compared to combining the inputs and feature extraction stages in a monolithic framework. Specifically, if one input mode suffers from obviously degraded quality, DME can still use the other mode to perform recognition. We highlight this flexibility in Section~\ref{QE}.

After $X_{f}$ and $X_{s}$ are extracted, they pass through the respective MLP networks to extract the video identification embeddings $l_{f}$ and $l_{s}$, which are defined as,

\begin{equation}
    l_{f} = \mathcal{M}_{f}(X_{f}),\\
    l_{s} = \mathcal{M}_{s}(X_{s}),
\end{equation}
where $\mathcal{M}_{f}$ and $\mathcal{M}_{s}$ are the MLP networks. For identification, a triplet loss~\cite{DBLP:journals/corr/HofferA14} is used to maximize the distance of embeddings from different subjects, and minimize the distance from the same subject. Specifically:

\begin{equation}
\begin{split}
    \mathcal{L}^{tri}_{f} = [\gD(l_{f}(i),l_{f}(k))-\gD(l_{f}(i),l_{f}(j))+m]_{+},\\
    \mathcal{L}^{tri}_{s} = [\gD(l_{s}(i),l_{s}(k))-\gD(l_{s}(i),l_{s}(j))+m]_{+},\\
\end{split}
\end{equation}
where $i$ and $j$ are videos from the same subject, $k$ is the video from another subject. $\gD(*)$ is the Euclidean distance measure, $m$ is the margin of the triplet loss, and operation
$[*]_+$ describes $\max(*, 0)$. The overall loss is:

\begin{equation}
% \begin{split}
    \mathcal{L}^{tot} = \mathcal{L}^{tri}_{f}+\mathcal{L}^{tri}_{s}.
    %+\lambda_{CE}(\mathcal{L}^{CE}_{f}+\mathcal{L}^{CE}_{s})\\
% \end{split}
\end{equation}

At test time, DME obtains the \emph{ensemble feature} through concatenation, i.e. $l_i=l_{f}\oplus l_{s}$  from every video, and uses $l_i$ to perform recognition. We find this to be a simple yet effective strategy that produces more discriminative features than those from individual modalities. 

% This setup also allows $\mathcal{F}_{f}$ and $\mathcal{F}_{s}$ to focus on their individual modalities, and outperforms a joint-learning alternative, where $X_i$ is used at train time to calculate the triplet loss. 

In the following section, we expand on the design details of individual feature extraction components; in particular, we propose a novel GaitPattern framework.

% To get a robust motion feature from gait, we introduce a gait representation method, \emph{GaitPattern}, making use of temporal information. 

\input{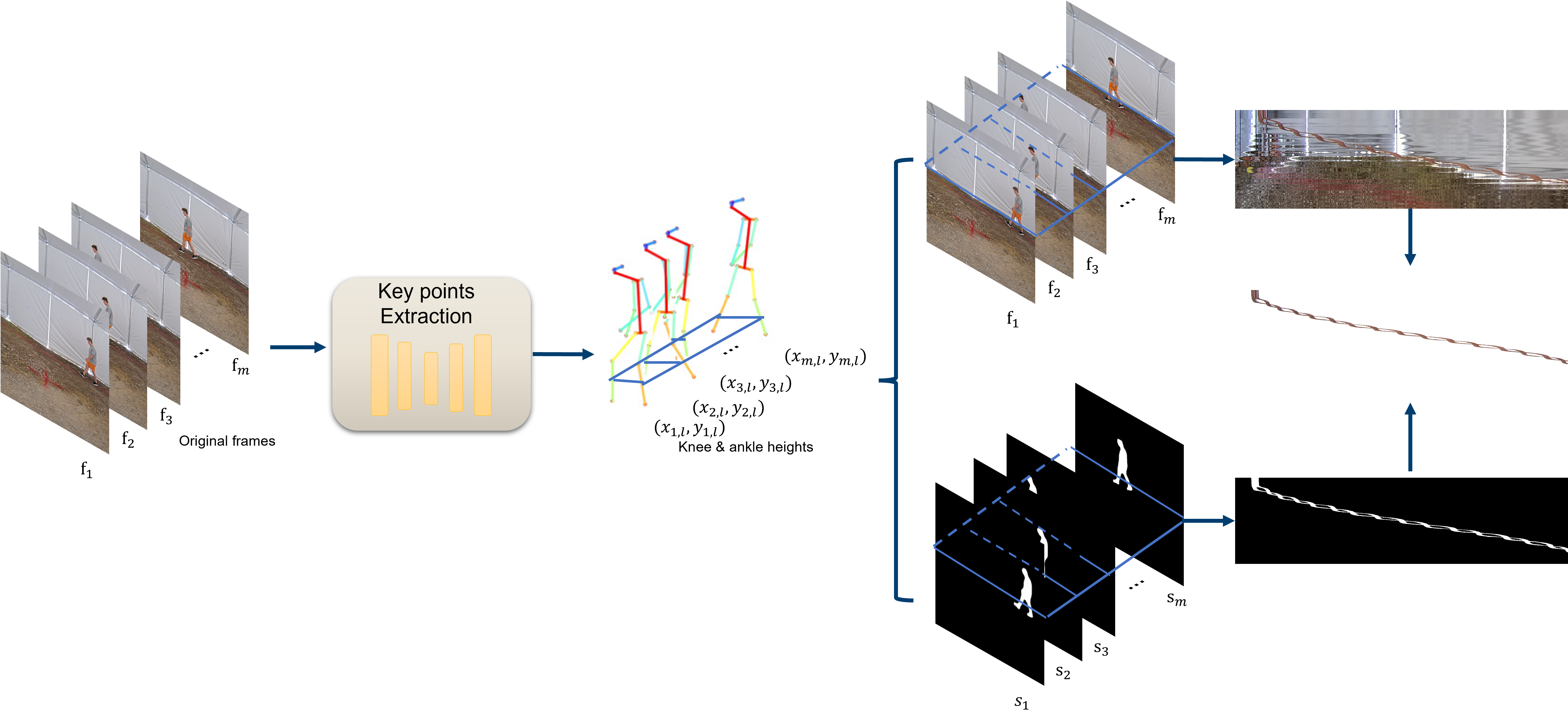}
\subsection{GaitPattern}\label{GFE}

A gait representation framework should focus on the walking motion and body shape and generate features robust to the identities' appearance. We propose GaitPattern, which incorporates both conventional silhouette-based feature extractor and an additional novel Double Helical Signature branch to better encode information like motion and viewing angles.

\subsubsection{Gait Feature Extraction}

A standard gait feature extraction model takes the silhouette sequences $\mS$ as input and generates the gait feature $X_{g}$. Previous works~\cite{fan2020gaitpart,chao2019gaitset,lin2021gait} mainly apply the "CL-SP-CL-SP-CL" pattern feature extractor in which CL represents a convolutional layer and SP, a spatial pooling layer. They use different kinds of convolution, i.e. 2D and 3D convolution, etc., and structures, i.e., global, local features combination and feature map partition, etc., to generate features. Then a temporal pooling step is applied to shrink the time dimension to extract a same-sized feature. This process is described as follows:
\begin{equation}
    X_{g} = \mathcal{F}_{s}(S) = \underset{t}{\max}(\mathcal{E}_{g}(\mS)),
\end{equation}
where $\mathcal{E}_{g}$ is a gait feature extractor and $\underset{t}{\max}$ represents temporal pooling. Our gait feature extraction model could flexibly apply the recent gait recognition methods as a backbone. By introducing the temporal pooling operation $\underset{t}{\max}$, the temporal information is lost, so we propose a different method to increase the temporal resolution.

\subsubsection{DHS Feature Extraction}
Based on previous works, it appears that gait recognition performance is reduced due to information loss caused by temporal pooling and different viewing angles. Based on our analysis, we desire to preserve some information in the time domain to make up for the information loss caused by temporal pooling. Then it is crucial to provide viewpoint information and regularize the model to generate robust features. We observe that the DHS contains this information as it records the dynamic movement of knees in 2D. The knee movement is representative and a distinguishable feature of human gait. The DHS patterns as a function of viewing angles are shown in Fig.~\ref{fig:three graphs}. We embed the viewing angle information into the DHS to help the model generate robust features for the same identity in different angles, which mitigates the effects of feature variations at different view points.

\input{FG2023/pictures/samples}

As mentioned in \ref{subsec:DHS}, DHS is an efficient representation that describes many gait attributes. In the past, DHS had been estimated at fixed view points and known location of the subjects' knees. We improve this feature extraction process with a CNN-based key-point extraction model and obtain the knee heights $\mathcal{H}_{\textrm{knee}}(t)\in \mathbb{R}$ in every frame $\vf_i$. Along with the silhouette mask $\vs_i$, this allows us to obtain the RGB width-time image $\mI_{\textrm{knee}}$, its mask $\vs_{\textrm{knee}}$, and subsequently the DHS $g_{\textrm{DHS}}$. As shown in Fig.~\ref{fig:sub1}, this process can be described as:

\begin{equation}
\begin{split}
  \mI_{\textrm{knee}}(x,t) = \vf_t(x,\mathcal{H}_{\textrm{knee}}(t)),\\
    % \mI_{\textrm{knee}}(x,t) = \mV(x,\mathcal{H}_{\textrm{knee}}(t),t),\\
  \vs_{\textrm{knee}}(x,t) = \vs_t (x,\mathcal{H}_{\textrm{knee}}(t)),\\
  g_{\textrm{DHS}} = \mI_{\textrm{knee}}\odot \vs_{\textrm{knee}}.
\end{split}
\end{equation}

After the DHS is generated, we extract the appropriate features from it and incorporate into a modern gait recognition architecture. In this process, one key consideration is the duration of the video. Since we want to increase the temporal resolution, we do not apply temporal pooling which is widely used to deal with varying temporal length. Based on the Ground Principle of Gait Temporal Aggregation \cite{fan2020gaitpart}, to make the system robust, the feature from the sequence containing at least a complete cycle should be the same for periodic motion under ideal conditions. Denote $ {g_{\textrm{DHS}}}(x,1:m)$ as the DHS stacked from frame $\vf_1$ to $\vf_m$. So the feature extractor for DHS should satisfy the following condition:
\begin{equation}
\begin{aligned}
    \mathcal{F}_{d}({g_{\textrm{DHS}}}(x,s:e)) &= \mathcal{F}_{d}({g_{\textrm{DHS}}}(x,s:s+C)), \quad
    \forall e-s > C
\end{aligned}
\end{equation}
\noindent where $\mathcal{F}_{d}$ is the sequence feature extraction module, $C$ is the complete cycle of the motion, $s$ and $e$ are selected start and end time (height in $g_{\textrm{DHS}}$), $s, e \in \{1,2,\dots,m\}$.

To use the previous principle, we apply $\max(\cdot)$ to select the DHS feature to be fused latter. Therefore, the DHS feature extraction process is designed as follows:
\begin{equation}
\begin{aligned}
    \mX_{DHS} = \max(&\mathcal{F}_{d}({g_{\textrm{DHS}}}(x,s_1:e_1)), 
    \\&\dots, 
    \\&\mathcal{F}_{d}({g_{\textrm{DHS}}}(x,s_n:e_n))).
\end{aligned}
\end{equation}
So the whole DHS is initially split into same-height intervals and fed to the sequence feature extraction module $F_{d}$. The final DHS feature $\mX_{DHS}$ is the largest one among all intervals' features. Compared to temporal pooling, the DHS feature extraction has the whole view of a video and maintains the temporal resolution with $e-s$ length.

We extract the DHS feature only when the videos are taken in a consistent environment. Since the key-point estimation is accurate in a constrained situation, it makes the extracted DHS reliable. Therefore, for the indoor case, we just adjust the shape of the DHS feature to fit the size of the one from any gait recognition backbones, generating the silhouette modality feature and for outdoor, we directly apply the gait feature. The silhouette feature $X_{s}$ is as following:
\begin{equation}
    X_{s}= 
    \begin{cases}
      X_{g}\oplus X_{DHS}, & \text{indoor,} \\
      X_{g}, & \text{outdoor}.
    \end{cases}
\end{equation}

For unconstrained scenarios, only the silhouette feature is used. The silhouette feature $X_{s}$ is robust to the appearance change, improving the robustness of ensemble feature $l_i$.

So the GaitPattern is a flexible module which can shift among various gait recognition frameworks, and improves the overall recognition performance across viewing angles.

\subsection{RGB Feature Extraction}\label{rgbfeat}
For RGB feature extraction, we use a feature extractor similar to gait's feature extractor, but use the video $\mV$ as the input, i.e. the RGB feature $X_{f}$ is
\begin{equation}
    X_{f} = (\underset{t}{\max}(\mathcal{E}_{f}(\mV))),
\end{equation}
in which $\mathcal{E}_{f}$ is a RGB feature extractor. RGB features $X_{f}$ provide rich information to enhance the appearance representation, and the feature backbones from~\cite{fan2020gaitpart,chao2019gaitset,lin2021gait} have shown to be efficient and effective architectures for video feature extraction.

%% file: FG2023/pictures/pattern.tex
\begin{figure}[!htb]
    \setlength{\abovecaptionskip}{3pt}
    \setlength{\tabcolsep}{2pt}
    \begin{tabular}[b]{c}
        \begin{subfigure}[b]{\linewidth}
            \includegraphics[width=\textwidth, height=.35\textwidth]{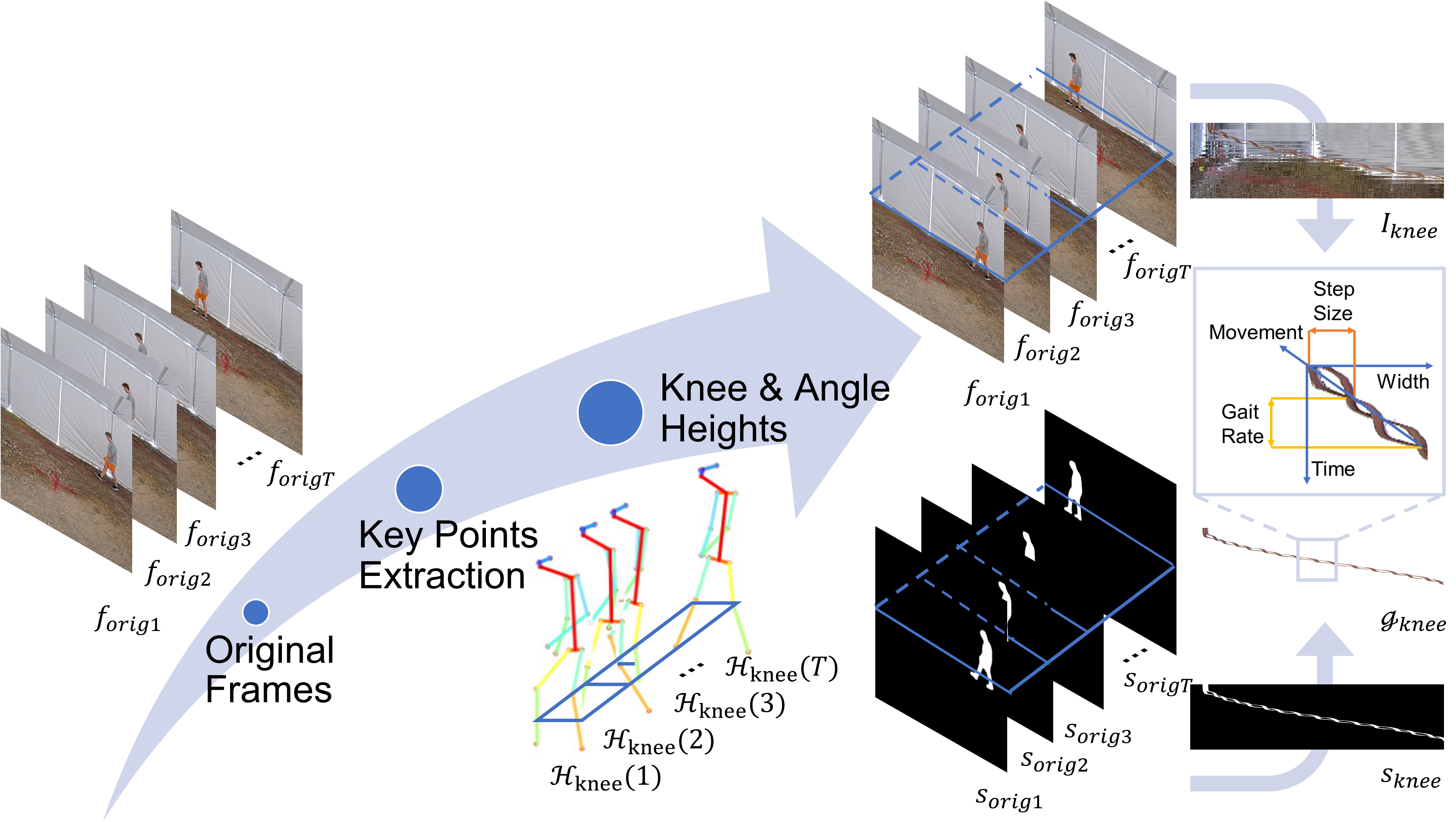}

            \label{fig:GaitPatterngeneraition}
        \end{subfigure}
    \end{tabular}
    \caption{The process of DHS generation.}
    \label{fig:sub1}
\end{figure}

% \begin{figure}[!htb]
%     \setlength{\abovecaptionskip}{3pt}
%     \setlength{\tabcolsep}{2pt}
%     \begin{tabular}[b]{cc}
%         \multicolumn{2}{c}{
%             \begin{subfigure}[b]{\linewidth}
%                 \includegraphics[width=\textwidth]{FG2023/pictures/Figure1_Version3.png}
%                 \caption{The process of GaitPattern generation}
%                 \label{fig:GaitPattern}
%             \end{subfigure}
%             } \\

% % \begin{figure}[!htb]
% %     \centering
% %     \includegraphics[width=\linewidth]{FG2023/pictures/pattern.png}
% %     \caption{The process of GaitPattern generation}
% %     \label{fig:GaitPattern}
% % \end{figure}    
    
%         \begin{subfigure}[b]{.48\linewidth}
%             \includegraphics[width=\textwidth,height=.8\textwidth]{FG2023/pictures/Figure2.png}
%             \caption{}
%             \label{bodychip}
%         \end{subfigure} &
%         \begin{subfigure}[b]{.48\linewidth}
%             \includegraphics[width=\textwidth,height=.8\textwidth]{FG2023/pictures/FIgure3.png}
%             \caption{}
%             \label{GaitPattern}
%         \end{subfigure} \\
%     \end{tabular}
%     \caption{The gait information embedded in frieze pattern on normalized silhouette~(\ref{bodychip}) and DHS.~(\ref{GaitPattern})}
%     \label{fig:sub1}
% \end{figure}

%% file: FG2023/pictures/samples.tex
% \begin{figure}[!htb]
%      \centering
%      \begin{subfigure}[b]{0.2\textwidth}
%          \centering
%          \includegraphics[width=\textwidth]{FG2023/pictures/CASIA-DHS/18.png}
%          \caption{$18\degree$}
%          \label{fig:y equals x}
%      \end{subfigure}
%      \hfill
%       \begin{subfigure}[b]{0.2\textwidth}
%          \centering
%          \includegraphics[width=\textwidth]{FG2023/pictures/CASIA-DHS/54.png}
%          \caption{$54\degree$}
%          \label{fig:y equals x}
%      \end{subfigure}
%      \hfill
%      \begin{subfigure}[b]{0.2\textwidth}
%          \centering
%          \includegraphics[width=\textwidth]{FG2023/pictures/CASIA-DHS/90.png}
%          \caption{$904\degree$}
%          \label{fig:three sin x}
%      \end{subfigure}
%      \hfill
%      \begin{subfigure}[b]{0.2\textwidth}
%          \centering
%          \includegraphics[width=\textwidth]{FG2023/pictures/CASIA-DHS/144.png}
%          \caption{$104\degree$}
%          \label{fig:five over x}
%      \end{subfigure}
%         \caption{The GaitPattern generated for CASIA-B at various viewing angles}
%         \label{fig:three graphs}
% \end{figure}

\begin{figure}[!htb]
    \setlength{\abovecaptionskip}{3pt}
    \setlength{\tabcolsep}{2pt}
    \centering
    \begin{tabular}[b]{cc}
        \begin{subfigure}[b]{.48\linewidth}
            \includegraphics[width=\textwidth,height=0.25\textwidth]{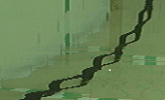}
            \caption{$18\degree$}
            \label{18degree}
        \end{subfigure} &
        \begin{subfigure}[b]{.48\linewidth}
            \includegraphics[width=\textwidth,height=0.25\textwidth]{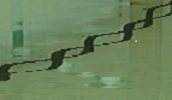}
            \caption{$54\degree$}
            \label{54degree}
        \end{subfigure} \\
        \begin{subfigure}[b]{.48\linewidth}
            \includegraphics[width=\textwidth,height=0.25\textwidth]{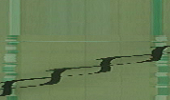}
            \caption{$90\degree$}
            \label{90degree}
        \end{subfigure} &
        \begin{subfigure}[b]{.48\linewidth}
            \includegraphics[width=\textwidth,height=0.25\textwidth]{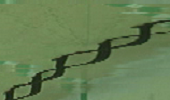}
            \caption{$104\degree$}
            \label{144degree}
        \end{subfigure} \\
    \end{tabular}
    \caption{The DHS generated for CASIA-B at various viewing angles.}
    \label{fig:three graphs}
\end{figure}

%\begin{figure}[!htb]
%      \centering
%      \begin{subfigure}[b]{0.34\textwidth}
%          \centering
%          \includegraphics[width=\textwidth]{FG2023/pictures/18.png}
%          \caption{$18\degree$}
%          \label{fig:y equals x}
%      \end{subfigure}
%      \hfill
%      \begin{subfigure}[b]{0.34\textwidth}
%          \centering
%          \includegraphics[width=\textwidth]{FG2023/pictures/54.png}
%          \caption{$54\degree$}
%          \label{fig:three sin x}
%      \end{subfigure}
%      \hfill
%      \begin{subfigure}[b]{0.34\textwidth}
%          \centering
%          \includegraphics[width=\textwidth]{FG2023/pictures/90.png}
%          \caption{$90\degree$}
%          \label{fig:five over x}
%      \end{subfigure}
%         \caption{The GaitPattern generated for CASIA-B at various viewing angles}
%         \label{fig:three graphs}
% \end{figure}

%% file: FG2023/experiments.tex
\section{Experiments}

\subsection{Datasets}
To evaluate the performance of our method, we perform extensive experiments on both controlled and unconstrained datasets. Specifically, we test on CASIA-B~\cite{yu2006framework}, which is acquired in a controlled, indoor environment. To further evaluate the robustness of whole body recognition in unconstrained scenarios, we test on the BRIAR dataset.

\input{FG2023/table/casia}
\textbf{CASIA-B~\cite{yu2006framework}} composes of 124 subjects, each of which has ten recorded sequences. In these ten sequences, six involve normal walking (NM), two involve walking with a bag (BG), and the rest are walking in a coat or jacket (CL). For each sequence, there are eleven videos recorded from viewing points ranging from  $0 \degree$ to $180 \degree$ at $18 \degree$ interval. All the videos are recorded in a constrained manner, which means that the videos are recorded indoor with consistent quality. The dataset is divided into training and testing sets following the popular protocol described in \cite{wu2016comprehensive}: first 74 identities are used for training and rest are for testing. During the training process, all sequences in the training set are fed to the model, i.e. NM, BG, and CL . For testing, the first four NM conditioned sequences (NM\#01-NM\#04) serve as gallery and the remaining sequence serve as the three probes, i.e. NM\#05-NM\#06, BG\#01-BG\#02 and CL\#01-CL\#02. This is to evaluate the robustness of the system on different walking types and dressing conditions.

\textbf{BRIAR} To evaluate the effectiveness of the proposed approach in unconstrained situations, we experimented with the BRIAR dataset. It includes 158 and 85 subjects serving as training and testing sets. Each identity has indoor and outdoor sequences, namely controlled and field sets. In the controlled set, there are ten cameras simultaneously recording the walking, and there are two walking types, i.e. structure walking and random walking. As for the field set, the videos are recorded at varying distances, i.e. close range, 100m, 200m, 400m, 500m and unmanned aerial vehicle (UAV). Except for the close range set, there is one viewing angle for each distance. And three cameras are applied at close range set in the same position but different yaw angles, i.e. $0 \degree$, $30 \degree$ and $50 \degree$. For all sequences, there are two clothing settings, i.e. set1 and set2 and in addition to the two walking types in controlled set, there are standing videos. Some examples are shown in Fig.~\ref{fig:vis}. For each video, the duration is about 90 seconds. And we get corresponding silhouettes using Detectron2~\cite{wu2019detectron2}.

We do the unconstrained gait recognition using the following protocol: In the training stage, the videos containing random and structure walking in the training set are fed to the model. In the test stage, we use `set1' videos in the controlled set as a gallery and `set2' videos containing gait, i.e. structure and random walking, in the field set as a probe. Typical examples of clothing changes are shown in Fig.~\ref{fig:vis} second and third columns. The length of the test videos ranges from 5 to 15 seconds.

\input{FG2023/pictures/unconstrained_vis}

\subsection{Implementation Details}
All the implementations are in Pytorch using four Nvidia A5000 GPUs. We adopt the popular preprocessing method~\cite{chao2019gaitset} to extract the gait silhouettes and RGB body chip for CASIA-B and BRIAR. The size for each frame is $64\times44$. To demonstrate the effectiveness of the proposed modules, GaitPattern and Dual-Model Ensemble, for CASIA-B, we follow the same settings as GaitSet~\cite{chao2019gaitset}, GaitPart~\cite{fan2020gaitpart} and GaitGL~\cite{lin2021gait} for gait and body feature extraction. We use Resnet18 as the DHS feature extraction backbone. For the BRIAR dataset, on account of the complexity and duration of the videos, we enlarge the total iterations to 120k for three models with (8,16) for the batch size, where the first digit is the number of subjects per batch and the second one is the number of videos per subject. All the backbones are trained with randomly sampled thirty frames from a video in each iteration and the selected 5 to 15 seconds sequences are used in the test phase.

\input{FG2023/table/briar_pro}

\subsection{Quantitative Evaluation}\label{QE}

\textbf{Evaluation on CASIA-B}~\cite{yu2006framework} We compare the proposed approach using GaitSet~\cite{chao2019gaitset}, GaitPart~\cite{fan2020gaitpart} and GaitGL~\cite{lin2021gait} with their own original methods on CASIA-B. The Rank-1 accuracy (\%) is shown in TABLE \ref{tab:CASIA-B}. We see that the GaitPattern improves the overall accuracy and with lower variance. The DME module boost the performance to a higher level.

Compared to the three backbones, the GaitPattern-based method improves the mean accuracy with lower variance across viewing angles, especially for the CL condition, improving by $6.9\%$, $10.8\%$ and $4.9\%$, respectively. The fusion of GaitPattern and GaitPart improves performance by $2.1\%$, $4.1\%$, $10.8\%$ in NM, BG and CL conditions respectively, reaching $98.3\%$, $95.6\%$ and $89.5\%$.

When it comes to DME, we combine the features from silhouettes, DHS and RGB. With the rich information from RGB, there is still a big improvement from the GaitPattern-based methods. And DME with GaitGL as backbone outperforms the others. Compared to the original GaitGL, the increase is $1.8\%$, $4.5\%$ and $9.2\%$ for NM, BG and CL conditions. Even compared to the GaitPattern-based GaitGL, DME with GaitGL has $1.7\%$, $4.0\%$ and $4.3\%$ improvements.

\textbf{Evaluation on BRIAR} 
We also apply GaitSet~\cite{chao2019gaitset}, GaitPart~\cite{fan2020gaitpart} and GaitGL~\cite{lin2021gait} to the proposed method. The Rank-1 accuracy (\%) is shown in TABLE~\ref{tab:unconstrained_result}. DME with GaitPart achieves the best overall recognition performance, reaching $64.21\%$. And GaitSet and GaitPart both have performance improvements of $15.07\%$, and $6.86\%$ respectively with the help of DME. But their performance have significant increase compared to only applying the RGB branch by $14.13\%$, $26.05\%$ and $3.85\%$, respectively. This indicates that the silhouette features enhances the effectiveness of aggregated features. We observe that DME with GaitGL has relatively low performance as the 3D convolution module does not extract robust features from a long video, especially with temporal stride. Further, from the fifth column of Fig.~\ref{fig:vis}, we find that as only the part of body above the knee is observable, the gait recognition performance at 200m is lower than at other distances.

%Further, from Fig.~\ref{fig:vis} second and third columns, we could find that even though the subjects are wearing the same clothes, the appearance is a bit different due to the camera setting, which leads to the RGB feature not being that helpful. So its 100m performance is lower than other distances'.

\input{FG2023/pictures/unconstrained_dis}

To demonstrate the effectiveness of the RGB feature, we experiment with the clothing protocol. For the BRIAR dataset, in the test stage, we use `set1' and `set2' sequences in the controlled set as a gallery and similarly `set1' and `set2' sequences in the field set as a probe.

From TABLE~\ref{tab:clothes_result}, we see that silhouette features do not perform well due to information loss. But when we introduce DME, the RGB feature helps improve the embedding's performance. For all three models, the DME results are even better than the RGB ones, which shows that the dual modalities complement mutually. Fig. \ref{fig:dis} demonstrates this property using distances between a gallery and probes.
\input{FG2023/table/briar_nochange}

%% file: FG2023/table/casia.tex
\begin{table*}[!htb]
\centering
\begin{tabular}{| c | c |c |c |c |c |c |c |c |c |c |c |c |c |}
\hline
\multicolumn{2}{|c|}{Gallery NM\#1-4} & \multicolumn{12}{c|}{$0\degree - 180\degree$}\\
\hline
\multicolumn{2}{|c|}{Probe} & $0\degree$ & $18\degree$ & $36\degree$ & $54\degree$ & $72\degree$& $90\degree$& $108\degree$ & $126\degree$ & $144\degree$& $162\degree$& $180\degree$& mean \\
\hline
\multirow{7}{*}{NM\#5-6} & GaitSet~\cite{chao2019gaitset} & 90.8 & 97.9 & 99.4 & 96.9 & 93.6 & 91.7 & 95.0 & 97.8 & 98.9 & 96.8 & 85.8 & 95.0 \\\
 & GaitPart~\cite{fan2020gaitpart} & 94.1 & 98.6 & 99.3 & 98.5 &  94.0 & 92.3 & 95.9 & 98.4 & 99.2 & 97.8 & 90.4 & 96.2\\ 
 & GaitGL~\cite{lin2021gait} & 96.0 & 98.3 & 99.0 & 97.9 & 96.9 & 95.4 & 97.0 & 98.9 & 99.3 & 98.8 & 94.0 & 97.4 \\  
 & GaitPattern (w/~\cite{chao2019gaitset}) & 92.2 & 98.4 & 98.5 & 97.9 & 94.7
 & 92.5 & 95.6 & 97.1 & 98.3 & 98.2 & 88.4 & 95.6 \\ 
 & GaitPattern (w/~\cite{fan2020gaitpart}) & \underline{96.9} & 98.4 & \underline{99.1} & \underline{98.8} & \underline{98.7} & \underline{97.1} & \underline{98.1} & \underline{99.0} & \underline{99.4} & \underline{99.0} & \underline{97.3} & \underline{98.3} \\
 & GaitPattern (w/~\cite{lin2021gait}) & 96.1 & \underline{98.7} & 99.0 & 98.1 & 95.6 & 96.5 & 97.6 & 98.3 & 98.8 & 98.7 & 95.1 & 97.5 \\
 \cline{2-14}
 & DME (w/~\cite{fan2020gaitpart}) & \textbf{99.2} & 99.0 & 99.2 & 98.9 & \textbf{99.2} & 98.6 & 99.0 & 99.3 & 99.4 & 99.2 & 99.4 & 99.1 \\
  & DME (w/~\cite{lin2021gait}) & 99.1 & \textbf{99.2} & \textbf{99.2} & \textbf{99.1} & 99.1 & \textbf{98.9} & \textbf{99.1} & \textbf{99.4} & \textbf{99.4} & \textbf{99.4} & \textbf{99.4} & \textbf{99.2} \\ 
\hline
\hline
\multirow{8}{*}{BG\#1-2} & GaitSet~\cite{chao2019gaitset} & 83.8 & 91.2 & 91.8 & 88.8 & 83.3 & 81.0 & 84.1 & 90.0 & 92.2 & 94.4 & 79.0 & 87.2 \\  
 & GaitPart~\cite{fan2020gaitpart} & 89.1 & 94.8 & 96.7 & 95.1 &  88.3 & 84.9 & 89.0 & 93.5 & 96.1 & 93.8 & 85.8 & 91.5\\   
 & GaitGL~\cite{lin2021gait} & 92.6 & 96.6 & 96.8 & 95.5 & 93.5 & 89.3 & 92.2 & 96.5 & 98.2 & 96.9 & 91.5 & 94.5 \\   
 & GaitPattern (w/~\cite{chao2019gaitset}) & 87.2 & 93.3 & 95.2 & 94.2 & 88.0
 & 82.5 & 85.8 & 92.8 & 96.7 & 95.4 & 85.2 & 90.6 \\   
 & GaitPattern (w/~\cite{fan2020gaitpart}) & \underline{94.2} & \underline{97.2} & 97.7 & \underline{96.9} & \underline{93.6} & \underline{91.9} & 94.7 & \underline{96.1} & \underline{98.1} & 97.4 & \underline{94.0} & \underline{95.6} \\   
 & GaitPattern (w/~\cite{lin2021gait}) & 92.8 & 97.2 & \underline{98.4} & 96.7 & 91.4 & 90.9 & \underline{95.0} & 95.5 & 97.3 & \underline{97.5} & 92.8 & 95.0 \\   
 \cline{2-14}
 & DME (w/~\cite{fan2020gaitpart}) & 98.8 & 99.0 & \textbf{99.0} & 98.8 & 97.9 & 97.2 & 98.1 & 99.0 & \textbf{99.4} & 99.2 & 98.9 & 98.7 \\
  & DME (w/~\cite{lin2021gait}) & \textbf{99.2} & \textbf{99.2} & 98.9 & \textbf{98.9} & \textbf{98.5} & \textbf{98.5} & \textbf{98.9} & \textbf{99.3} & 99.3 & \textbf{99.4} & \textbf{99.2} & \textbf{99.0} \\ 
\hline
\hline
\multirow{8}{*}{CL\#1-2} & GaitSet~\cite{chao2019gaitset} & 61.4 & 75.4 & 80.7 & 77.3 & 72.1 & 70.1 & 71.5 & 73.5 & 73.5 & 68.4 & 50.0 & 70.4 \\  
 & GaitPart~\cite{fan2020gaitpart} & 70.7 & 85.5 & 86.9 & 83.3 &  77.1 & 72.5 & 76.9 & 82.2 & 83.8 & 80.2 & 66.5 & 78.7\\   
 & GaitGL~\cite{lin2021gait} & 76.6 & 90.0 & 90.3 & 87.1 & 84.5 & 79.0 & 84.1 & 87.0 & 87.3 & 84.4 & 69.5 & 83.6 \\   
 & GaitPattern (w/~\cite{chao2019gaitset}) & 70.2 & 84.3 & 84.5 & 81.3 & 74.5
 & 72.5 & 74.0 & 79.0 & 83.0 & 80.2 & 66.5 & 77.3 \\   
 & GaitPattern (w/~\cite{fan2020gaitpart}) & \underline{85.0} & 92.5 & 93.4 & \underline{92.5} & \underline{88.9} & 83.7 & \underline{87.0} & \underline{90.7} & \underline{93.0} & \underline{90.5} & \underline{87.6} & \underline{89.5} \\   
 & GaitPattern (w/~\cite{lin2021gait}) & 83.4 & \underline{93.0} & \underline{94.6} & 90.8 & 85.4 & \underline{85.0} & 87.0 & 89.8 & 90.9 & 89.9 & 84.3 & 88.5 \\   
 \cline{2-14}
  & DME (w/~\cite{fan2020gaitpart}) & \textbf{92.3} & 95.9 & 95.9 & 94.3 & \textbf{92.2} & 88.5 & 88.3 & 93.2 & 93.7 & 90.8 & 86.6 & 92.0 \\
  & DME (w/~\cite{lin2021gait}) & 92.1 & \textbf{96.0} & \textbf{96.6} & \textbf{95.0} & 91.1 & \textbf{90.3} & \textbf{92.0} & \textbf{93.2} & \textbf{93.8} & \textbf{92.6} & \textbf{88.7} & \textbf{92.8} \\ 
\hline
\end{tabular}
\caption{Rank-1 accuracy (\%) on CASIA-B  excluding identical-view case. }
\label{tab:CASIA-B}
\end{table*}

%% file: FG2023/pictures/unconstrained_vis.tex
\begin{figure}[!htb]
    \setlength{\abovecaptionskip}{3pt}
    \setlength{\tabcolsep}{0pt}
    \renewcommand\arraystretch{0}
    \setlength{\belowcaptionskip}{-0.2cm}
    \centering
    \begin{tabular}[c]{cccccccc}
    
        \begin{subfigure}[b]{.12\linewidth}
             \includegraphics[width=\textwidth,height=1.5\textwidth]{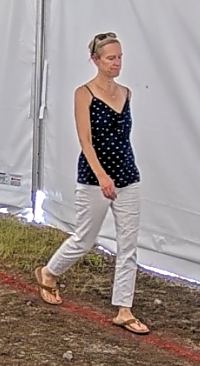}
                \label{fig:265controlled}
        \end{subfigure}&
        \begin{subfigure}[b]{.12\linewidth}
             \includegraphics[width=\textwidth,height=1.5\textwidth]{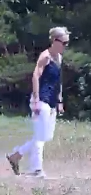}
                \label{fig:265_100}
        \end{subfigure}&
        \begin{subfigure}[b]{.12\linewidth}
             \includegraphics[width=\textwidth,height=1.5\textwidth]{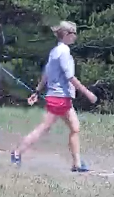}
                \label{fig:265_100_set2}
        \end{subfigure}&
        \begin{subfigure}[b]{.12\linewidth}
             \includegraphics[width=\textwidth,height=1.5\textwidth]{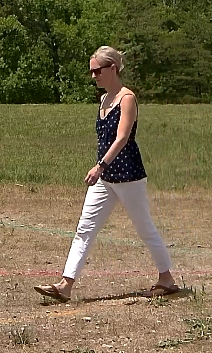}
                \label{fig:265_close}
        \end{subfigure}&
        \begin{subfigure}[b]{.12\linewidth}
             \includegraphics[width=\textwidth,height=1.5\textwidth]{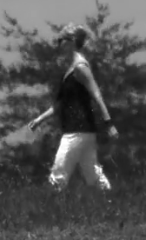}
                \label{fig:265_200}
        \end{subfigure}&
        \begin{subfigure}[b]{.12\linewidth}
             \includegraphics[width=\textwidth,height=1.5\textwidth]{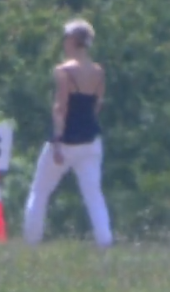}
                \label{fig:265_400}
        \end{subfigure}&
        \begin{subfigure}[b]{.12\linewidth}
             \includegraphics[width=\textwidth,height=1.5\textwidth]{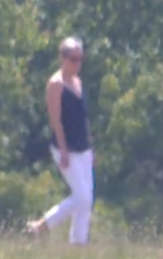}
                \label{fig:265_500}
        \end{subfigure}&
        \begin{subfigure}[b]{.12\linewidth}
             \includegraphics[width=\textwidth,height=1.5\textwidth]{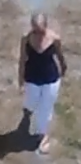}
                \label{fig:265_uav}
        \end{subfigure}
        \vspace{-1.2em} \\
        \begin{subfigure}[b]{.12\linewidth}
             \includegraphics[width=\textwidth,height=1.5\textwidth]{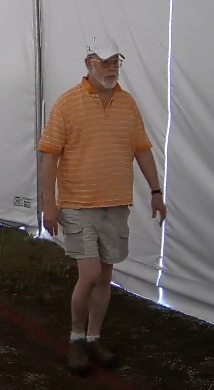}
                \label{fig:266_controlled}
        \end{subfigure}&
        \begin{subfigure}[b]{.12\linewidth}
             \includegraphics[width=\textwidth,height=1.5\textwidth]{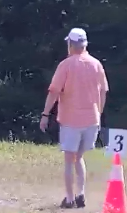}
                \label{fig:266_100}
        \end{subfigure}&
        \begin{subfigure}[b]{.12\linewidth}
             \includegraphics[width=\textwidth,height=1.5\textwidth]{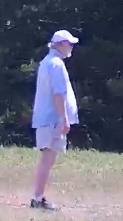}
                \label{fig:266_100_set}
        \end{subfigure}&
        \begin{subfigure}[b]{.12\linewidth}
             \includegraphics[width=\textwidth,height=1.5\textwidth]{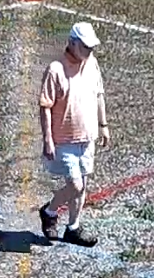}
                \label{fig:266_close}
        \end{subfigure}&
        \begin{subfigure}[b]{.12\linewidth}
             \includegraphics[width=\textwidth,height=1.5\textwidth]{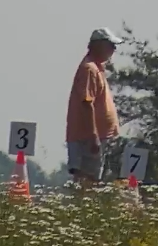}
                \label{fig:266_200}
        \end{subfigure}&
        \begin{subfigure}[b]{.12\linewidth}
             \includegraphics[width=\textwidth,height=1.5\textwidth]{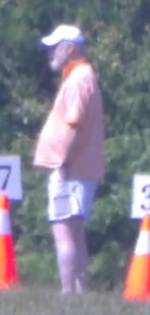}
                \label{fig:266_400}
        \end{subfigure}&
        \begin{subfigure}[b]{.12\linewidth}
             \includegraphics[width=\textwidth,height=1.5\textwidth]{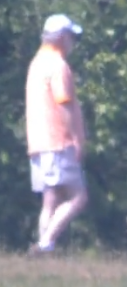}
                \label{fig:266_500}
        \end{subfigure}&
        \begin{subfigure}[b]{.12\linewidth}
             \includegraphics[width=\textwidth,height=1.5\textwidth]{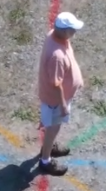}
                \label{fig:266_uav}
        \end{subfigure}
        \vspace{-1.2em}\\
        \begin{subfigure}[b]{.12\linewidth}
             \includegraphics[width=\textwidth,height=1.5\textwidth]{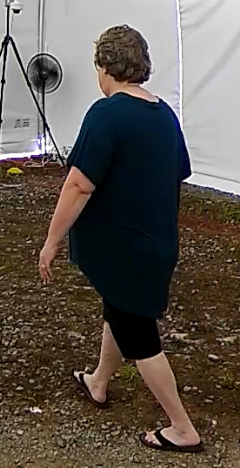}
                \label{fig:267_controlled}
        \end{subfigure}&
        \begin{subfigure}[b]{.12\linewidth}
             \includegraphics[width=\textwidth,height=1.5\textwidth]{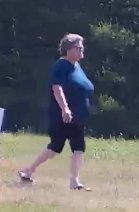}
                \label{fig:267_100}
        \end{subfigure}&
        \begin{subfigure}[b]{.12\linewidth}
             \includegraphics[width=\textwidth,height=1.5\textwidth]{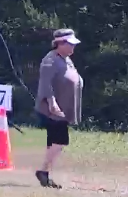}
                \label{fig:267_100_set}
        \end{subfigure}&
        \begin{subfigure}[b]{.12\linewidth}
             \includegraphics[width=\textwidth,height=1.5\textwidth]{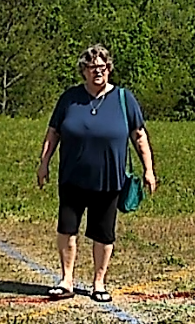}
                \label{fig:267_close}
        \end{subfigure}&
        \begin{subfigure}[b]{.12\linewidth}
             \includegraphics[width=\textwidth,height=1.5\textwidth]{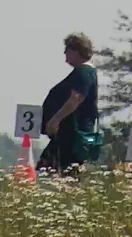}
                \label{fig:267_200}
        \end{subfigure}&
        \begin{subfigure}[b]{.12\linewidth}
             \includegraphics[width=\textwidth,height=1.5\textwidth]{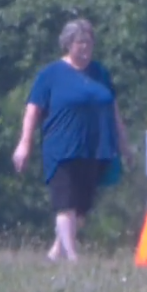}
                \label{fig:267_400}
        \end{subfigure}&
        \begin{subfigure}[b]{.12\linewidth}
             \includegraphics[width=\textwidth,height=1.5\textwidth]{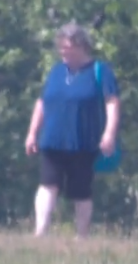}
                \label{fig:267_500}
        \end{subfigure}&
        \begin{subfigure}[b]{.12\linewidth}
             \includegraphics[width=\textwidth,height=1.5\textwidth]{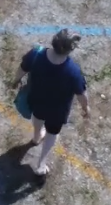}
                \label{fig:267_uav}
        \end{subfigure}
        \vspace{-1.2em}\\
        \begin{subfigure}[b]{.12\linewidth}
             \includegraphics[width=\textwidth,height=1.5\textwidth]{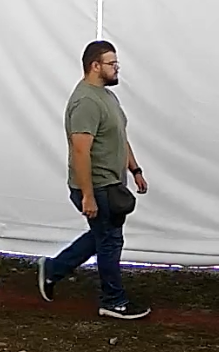}
                \label{fig:289_controlled}
        \end{subfigure}&
        \begin{subfigure}[b]{.12\linewidth}
             \includegraphics[width=\textwidth,height=1.5\textwidth]{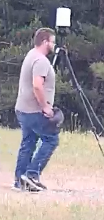}
                \label{fig:289_100}
        \end{subfigure}&
        \begin{subfigure}[b]{.12\linewidth}
             \includegraphics[width=\textwidth,height=1.5\textwidth]{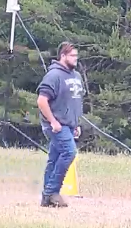}
                \label{fig:289_100_set}
        \end{subfigure}&
        \begin{subfigure}[b]{.12\linewidth}
             \includegraphics[width=\textwidth,height=1.5\textwidth]{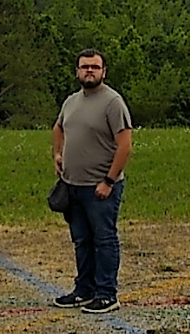}
                \label{fig:289_close}
        \end{subfigure}&
        \begin{subfigure}[b]{.12\linewidth}
             \includegraphics[width=\textwidth,height=1.5\textwidth]{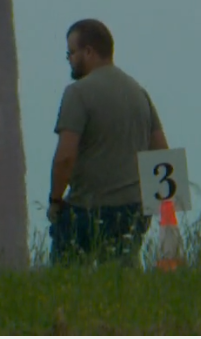}
                \label{fig:289_200}
        \end{subfigure}&
        \begin{subfigure}[b]{.12\linewidth}
             \includegraphics[width=\textwidth,height=1.5\textwidth]{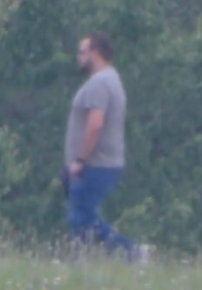}
                \label{fig:289_400}
        \end{subfigure}&
        \begin{subfigure}[b]{.12\linewidth}
             \includegraphics[width=\textwidth,height=1.5\textwidth]{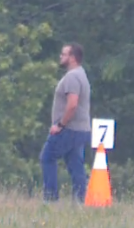}
                \label{fig:289_500}
        \end{subfigure}&
        \begin{subfigure}[b]{.12\linewidth}
             \includegraphics[width=\textwidth,height=1.5\textwidth]{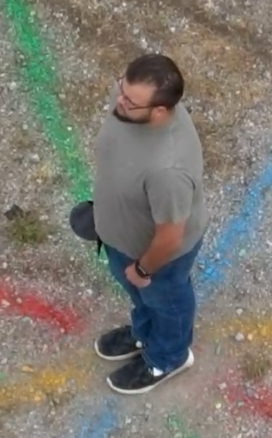}
                \label{fig:289_uav}
        \end{subfigure}
        \vspace{-1.2em}\\
    \end{tabular}
    \caption{Examples of a subject in different conditions from the BRIAR dataset at different distances and clothing.The columns represents controlled, 100m-set1, 100m-set2, close range, 200m, 400m, 500m and UAV respectively.}
    \label{fig:vis}
\end{figure}

%% file: FG2023/table/briar_pro.tex
\begin{table}[h]
\setlength{\belowcaptionskip}{-0.3cm}
\setlength{\tabcolsep}{3pt}
\centering
\begin{tabular}{ | c |c |c |c |c  |c |c | c |}
\hline
\multicolumn{1}{|c|}{Probe} & Close range & 100m  &200m& 400m & 500m & UAV& Mean  \\
\hline
GaitSet & 55.40 & 56.13 & 30.47 & 46.76  & 44.98  & 65.52 & 49.87\\
\cline{2-8}
GaitPart & 60.56 & 71.09 & 35.62 & 50.78 & 50.19 & \textbf{75.86} & 57.35  \\
\cline{2-8}
GaitGL & 55.40 & 66.05 & 16.31 & 38.93  &40.89& 62.07 & 46.60 \\
\cline{2-8}
RGB (w/~\cite{chao2019gaitset}) &  50.23 & 57.65  & 37.34 & 48.77 & 39.78 & 37.93 & 45.26 \\
\cline{2-8}
RGB (w/~\cite{fan2020gaitpart})  & 48.04 & 45.71  & 19.96 & 35.12 & 35.32 & 44.83 & 38.16 \\
\cline{2-8}
RGB (w/~\cite{lin2021gait})  & 39.75& 45.55  &18.03 & 31.77 &24.16 & 24.14 & 30.56  \\
\cline{2-8}
DME (w/~\cite{chao2019gaitset}) &  67.92 & 71.76  & \textbf{44.21} & \textbf{64.65} &56.13 &51.72 & 59.39 \\
\cline{2-8}
DME (w/~\cite{fan2020gaitpart})  & \textbf{69.64} & \textbf{77.31}  & 43.35 &64.21 &\textbf{58.36} &72.41 & \textbf{64.21} \\
\cline{2-8}
DME (w/~\cite{lin2021gait})  & 43.04 & 51.43  & 19.96 &36.91 &27.88 &27.59 & 34.46  \\
\cline{2-8}

\hline
\end{tabular}
\caption{Rank-1 accuracy (\%) on BRIAR.}
\label{tab:unconstrained_result}
\end{table}

%% file: FG2023/pictures/unconstrained_dis.tex
\begin{figure}[!htb]
    \setlength{\abovecaptionskip}{3pt}
    \setlength{\tabcolsep}{2pt}
    \centering
    \begin{tabular}[c]{cccc}
    
     \begin{subfigure}[b]{.21\linewidth}
             \includegraphics[width=\textwidth,height=1.4\textwidth]{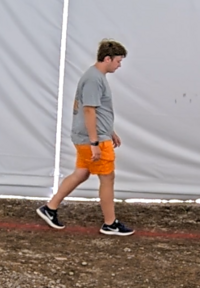}
                \caption{Gallery}
                \label{fig:controlled}
        \end{subfigure}&
        \begin{subfigure}[b]{.21\linewidth}
             \includegraphics[width=\textwidth,height=1.4\textwidth]{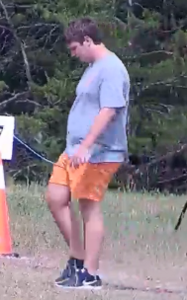}
                \caption{Probe}
                \label{fig:100m-set1}
        \end{subfigure}&
        \begin{subfigure}[b]{.21\linewidth}
             \includegraphics[width=\textwidth,height=1.4\textwidth]{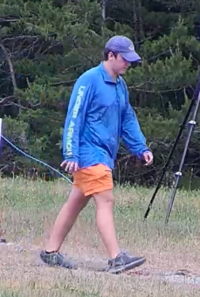}
                \caption{Probe}
                \label{fig:100m-set2}
        \end{subfigure}&
        \begin{subfigure}[b]{.21\linewidth}
             \includegraphics[width=\textwidth,height=1.4\textwidth]{FG2023/pictures/unconstrained_vis/265/265_100m_set1.png}
                \caption{Probe}
                \label{fig:closerange}
        \end{subfigure}\\
    \small{Gait}    & \small{0.84} & \small{0.85} & \small{1.18} \\
    \small{RGB}     & \small{0.75} & \small{0.98} & \small{1.05} \\
    \small{Dual}    & \small{0.81} & \small{0.91} & \small{1.11} \\

    \end{tabular}
    \caption{The Euclidean distances of probes to gallery for gait, RGB and dual features (GaitPart backbone).}
    % \vspace{-2em}
    \label{fig:dis}
\end{figure}

%% file: FG2023/table/briar_nochange.tex
\begin{table}[h]
\centering
\begin{tabular}{ | c |c|c |c|}
\hline
Backbone & Gait & RGB & Dual  \\
\hline
GaitSet~\cite{chao2019gaitset} & 86.31 & 92.53 & 95.89 \\
% \cline{2-3}
%  & RGB &\\
% \cline{2-3}
%  & Dual & \\
\hline
GaitPart~\cite{fan2020gaitpart}& 73.29& 88.27 & 94.19 \\
% \cline{2-3}
%  &RGB \\
% \cline{2-3}
%  & Dual  \\
\hline
GaitGL~\cite{lin2021gait}& 65.56& 86.29 & 87.36\\
% \cline{2-3}
% &RGB \\ 
% \cline{2-3}
% & Dual  \\

\hline
\end{tabular}
\caption{Mean accuracy ($\%$) on the privately collected unconstrained outdoor dataset with changing clothes.}
\label{tab:clothes_result}
\end{table}

%% file: FG2023/conclusion.tex
\section{Conclusion}

In this work, we present Dual-Modal Ensemble, which leverages both the human silhouette and RGB to perform whole-body recognition. Specifically, DME leverages the rich information in RGB image to boost the general recognition performance, while also maintaining high performance on challenging conditions, e.g. with changing clothes, based on a gait recognition sub-network. DME accomplishes this by training a powerful gait and body recognition network individually, and concatenating features produced by each modality to obtain the final feature of the video. DME shows very robust performance in unconstrained conditions, demonstrating the effectiveness of combining features from different modalities. Within DME, we propose a gait recognition network called GaitPattern, inspired by the traditional gait pattern analysis. GaitPattern generates and incorporates a gait representation called Double Helical Signature, and can be used alongside popular gait recognition architectures to complement the lack of view angle information and spatial resolution. We find that GaitPattern can significantly enhance performances of current SoTA methods, particularly over challenging view angles. Future work includes investigating more robust key-point detection algorithm such that GaitPattern can be better incorporated into unconstrained environments, as well as reducing the computational complexity of multi-modal ensemble by fusing models at early stages without losing discriminative power for each modality. 

%% file: FG2023/acknowledge.tex
\section{acknowledgement}

This research is based upon work supported in part by the Office of the Director of National Intelligence (ODNI), Intelligence Advanced Research Projects Activity (IARPA), via [2022-21102100005]. The views and conclusions contained herein are those of the authors and should not be interpreted as necessarily representing the official policies, either expressed or implied, of ODNI, IARPA, or the U.S. Government. The US. Government is authorized to reproduce and distribute reprints for governmental purposes notwithstanding any copyright annotation therein.

%% file: GaitPattern.bbl
\begin{thebibliography}{10}\itemsep=-1pt

\bibitem{an2020performance}
W.~An, S.~Yu, Y.~Makihara, X.~Wu, C.~Xu, Y.~Yu, R.~Liao, and Y.~Yagi.
\newblock Performance evaluation of model-based gait on multi-view very large
  population database with pose sequences.
\newblock {\em IEEE transactions on biometrics, behavior, and identity
  science}, 2(4):421--430, 2020.

\bibitem{ariyanto2011model}
G.~Ariyanto and M.~S. Nixon.
\newblock Model-based 3d gait biometrics.
\newblock In {\em 2011 international joint conference on biometrics (IJCB)},
  pages 1--7. IEEE, 2011.

\bibitem{bai2022salient}
S.~Bai, B.~Ma, H.~Chang, R.~Huang, and X.~Chen.
\newblock Salient-to-broad transition for video person re-identification.
\newblock In {\em Proceedings of the IEEE/CVF Conference on Computer Vision and
  Pattern Recognition}, pages 7339--7348, 2022.

\bibitem{bodor2009view}
R.~Bodor, A.~Drenner, D.~Fehr, O.~Masoud, and N.~Papanikolopoulos.
\newblock View-independent human motion classification using image-based
  reconstruction.
\newblock {\em Image and Vision Computing}, 27(8):1194--1206, 2009.

\bibitem{chao2019gaitset}
H.~Chao, Y.~He, J.~Zhang, and J.~Feng.
\newblock Gaitset: Regarding gait as a set for cross-view gait recognition.
\newblock In {\em Proceedings of the AAAI conference on artificial
  intelligence}, volume~33, pages 8126--8133, 2019.

\bibitem{fan2020gaitpart}
C.~Fan, Y.~Peng, C.~Cao, X.~Liu, S.~Hou, J.~Chi, Y.~Huang, Q.~Li, and Z.~He.
\newblock Gaitpart: Temporal part-based model for gait recognition.
\newblock In {\em Proceedings of the IEEE/CVF conference on computer vision and
  pattern recognition}, pages 14225--14233, 2020.

\bibitem{fu2019sta}
Y.~Fu, X.~Wang, Y.~Wei, and T.~Huang.
\newblock Sta: Spatial-temporal attention for large-scale video-based person
  re-identification.
\newblock In {\em Proceedings of the AAAI conference on artificial
  intelligence}, volume~33, pages 8287--8294, 2019.

\bibitem{han2005individual}
J.~Han and B.~Bhanu.
\newblock Individual recognition using gait energy image.
\newblock {\em IEEE transactions on pattern analysis and machine intelligence},
  28(2):316--322, 2005.

\bibitem{DBLP:journals/corr/HofferA14}
E.~Hoffer and N.~Ailon.
\newblock Deep metric learning using triplet network.
\newblock In Y.~Bengio and Y.~LeCun, editors, {\em 3rd International Conference
  on Learning Representations, {ICLR} 2015, San Diego, CA, USA, May 7-9, 2015,
  Workshop Track Proceedings}, 2015.

\bibitem{hou2020gait}
S.~Hou, C.~Cao, X.~Liu, and Y.~Huang.
\newblock Gait lateral network: Learning discriminative and compact
  representations for gait recognition.
\newblock In {\em European conference on computer vision}, pages 382--398.
  Springer, 2020.

\bibitem{huang20213d}
Z.~Huang, D.~Xue, X.~Shen, X.~Tian, H.~Li, J.~Huang, and X.-S. Hua.
\newblock 3d local convolutional neural networks for gait recognition.
\newblock In {\em Proceedings of the IEEE/CVF International Conference on
  Computer Vision}, pages 14920--14929, 2021.

\bibitem{kale2002gait}
A.~Kale, A.~Rajagopalan, N.~Cuntoor, and V.~Kruger.
\newblock Gait-based recognition of humans using continuous hmms.
\newblock In {\em Proceedings of Fifth IEEE International Conference on
  Automatic Face Gesture Recognition}, pages 336--341. IEEE, 2002.

\bibitem{kale2004identification}
A.~Kale, A.~Sundaresan, A.~Rajagopalan, N.~P. Cuntoor, A.~K. Roy-Chowdhury,
  V.~Kruger, and R.~Chellappa.
\newblock Identification of humans using gait.
\newblock {\em IEEE Transactions on image processing}, 13(9):1163--1173, 2004.

\bibitem{lau2021atfacegan}
C.~P. Lau, C.~D. Castillo, and R.~Chellappa.
\newblock Atfacegan: Single face semantic aware image restoration and
  recognition from atmospheric turbulence.
\newblock {\em IEEE Transactions on Biometrics, Behavior, and Identity
  Science}, 3(2):240--251, 2021.

\bibitem{lau2020atfacegan}
C.~P. Lau, H.~Souri, and R.~Chellappa.
\newblock Atfacegan: Single face image restoration and recognition from
  atmospheric turbulence.
\newblock In {\em 2020 15th IEEE International Conference on Automatic Face and
  Gesture Recognition (FG 2020)}, pages 32--39. IEEE, 2020.

\bibitem{li2019global}
J.~Li, J.~Wang, Q.~Tian, W.~Gao, and S.~Zhang.
\newblock Global-local temporal representations for video person
  re-identification.
\newblock In {\em Proceedings of the IEEE/CVF international conference on
  computer vision}, pages 3958--3967, 2019.

\bibitem{liao2020model}
R.~Liao, S.~Yu, W.~An, and Y.~Huang.
\newblock A model-based gait recognition method with body pose and human prior
  knowledge.
\newblock {\em Pattern Recognition}, 98:107069, 2020.

\bibitem{lin2021gait}
B.~Lin, S.~Zhang, and X.~Yu.
\newblock Gait recognition via effective global-local feature representation
  and local temporal aggregation.
\newblock In {\em Proceedings of the IEEE/CVF International Conference on
  Computer Vision}, pages 14648--14656, 2021.

\bibitem{liu2002gait}
Y.~Liu, R.~Collins, and Y.~Tsin.
\newblock Gait sequence analysis using frieze patterns.
\newblock In {\em European Conference on Computer Vision}, pages 657--671.
  Springer, 2002.

\bibitem{liu2004computational}
Y.~Liu, R.~T. Collins, and Y.~Tsin.
\newblock A computational model for periodic pattern perception based on frieze
  and wallpaper groups.
\newblock {\em IEEE transactions on pattern analysis and machine intelligence},
  26(3):354--371, 2004.

\bibitem{liu2006improved}
Z.~Liu and S.~Sarkar.
\newblock Improved gait recognition by gait dynamics normalization.
\newblock {\em IEEE Transactions on Pattern Analysis and Machine Intelligence},
  28(6):863--876, 2006.

\bibitem{niyogi1994analyzing}
S.~A. Niyogi and E.~H. Adelson.
\newblock Analyzing gait with spatiotemporal surfaces.
\newblock In {\em Proceedings of 1994 IEEE Workshop on Motion of Non-rigid and
  Articulated Objects}, pages 64--69. IEEE, 1994.

\bibitem{peng2021learning}
Y.~Peng, S.~Hou, K.~Ma, Y.~Zhang, Y.~Huang, and Z.~He.
\newblock Learning rich features for gait recognition by integrating skeletons
  and silhouettes.
\newblock {\em arXiv preprint arXiv:2110.13408}, 2021.

\bibitem{ran2010applications}
Y.~Ran, Q.~Zheng, R.~Chellappa, and T.~M. Strat.
\newblock Applications of a simple characterization of human gait in
  surveillance.
\newblock {\em IEEE Transactions on Systems, Man, and Cybernetics, Part B
  (Cybernetics)}, 40(4):1009--1020, 2010.

\bibitem{teepe2022towards}
T.~Teepe, J.~Gilg, F.~Herzog, S.~H{\"o}rmann, and G.~Rigoll.
\newblock Towards a deeper understanding of skeleton-based gait recognition.
\newblock In {\em Proceedings of the IEEE/CVF Conference on Computer Vision and
  Pattern Recognition}, pages 1569--1577, 2022.

\bibitem{teepe2021gaitgraph}
T.~Teepe, A.~Khan, J.~Gilg, F.~Herzog, S.~H{\"o}rmann, and G.~Rigoll.
\newblock Gaitgraph: Graph convolutional network for skeleton-based gait
  recognition.
\newblock In {\em 2021 IEEE International Conference on Image Processing
  (ICIP)}, pages 2314--2318. IEEE, 2021.

\bibitem{wang2022two}
L.~Wang and J.~Chen.
\newblock A two-branch neural network for gait recognition.
\newblock {\em arXiv preprint arXiv:2202.10645}, 2022.

\bibitem{wu2019detectron2}
Y.~Wu, A.~Kirillov, F.~Massa, W.-Y. Lo, and R.~Girshick.
\newblock Detectron2.
\newblock {https://github.com/facebookresearch/detectron2}, 2019.

\bibitem{wu2016comprehensive}
Z.~Wu, Y.~Huang, L.~Wang, X.~Wang, and T.~Tan.
\newblock A comprehensive study on cross-view gait based human identification
  with deep cnns.
\newblock {\em IEEE transactions on pattern analysis and machine intelligence},
  39(2):209--226, 2016.

\bibitem{yan2020learning}
Y.~Yan, J.~Qin, J.~Chen, L.~Liu, F.~Zhu, Y.~Tai, and L.~Shao.
\newblock Learning multi-granular hypergraphs for video-based person
  re-identification.
\newblock In {\em Proceedings of the IEEE/CVF conference on computer vision and
  pattern recognition}, pages 2899--2908, 2020.

\bibitem{yasarla2020learning}
R.~Yasarla and V.~M. Patel.
\newblock Learning to restore a single face image degraded by atmospheric
  turbulence using cnns.
\newblock {\em arXiv preprint arXiv:2007.08404}, 2020.

\bibitem{yasarla2021learning}
R.~Yasarla and V.~M. Patel.
\newblock Learning to restore images degraded by atmospheric turbulence using
  uncertainty.
\newblock In {\em 2021 IEEE International Conference on Image Processing
  (ICIP)}, pages 1694--1698. IEEE, 2021.

\bibitem{yu2006framework}
S.~Yu, D.~Tan, and T.~Tan.
\newblock A framework for evaluating the effect of view angle, clothing and
  carrying condition on gait recognition.
\newblock In {\em 18th International Conference on Pattern Recognition
  (ICPR'06)}, volume~4, pages 441--444. IEEE, 2006.

\bibitem{zhao20063d}
G.~Zhao, G.~Liu, H.~Li, and M.~Pietikainen.
\newblock 3d gait recognition using multiple cameras.
\newblock In {\em 7th International Conference on Automatic Face and Gesture
  Recognition (FGR06)}, pages 529--534. IEEE, 2006.

\bibitem{zhao2019attribute}
Y.~Zhao, X.~Shen, Z.~Jin, H.~Lu, and X.-s. Hua.
\newblock Attribute-driven feature disentangling and temporal aggregation for
  video person re-identification.
\newblock In {\em Proceedings of the IEEE/CVF conference on computer vision and
  pattern recognition}, pages 4913--4922, 2019.

\bibitem{zhou2002probabilistic}
S.~Zhou and R.~Chellappa.
\newblock Probabilistic human recognition from video.
\newblock In {\em European Conference on Computer Vision}, pages 681--697.
  Springer, 2002.

\bibitem{zhou2004visual}
S.~K. Zhou, R.~Chellappa, and B.~Moghaddam.
\newblock Visual tracking and recognition using appearance-adaptive models in
  particle filters.
\newblock {\em IEEE Transactions on Image Processing}, 13(11):1491--1506, 2004.

\bibitem{zhou2017see}
Z.~Zhou, Y.~Huang, W.~Wang, L.~Wang, and T.~Tan.
\newblock See the forest for the trees: Joint spatial and temporal recurrent
  neural networks for video-based person re-identification.
\newblock In {\em Proceedings of the IEEE conference on computer vision and
  pattern recognition}, pages 4747--4756, 2017.

\end{thebibliography}
